\documentclass{esannV2}
\usepackage[dvips]{graphicx}
\usepackage[latin1]{inputenc}
\usepackage{amssymb,amsmath,array}
\usepackage{comment}
\usepackage{url}

\usepackage{color}
\usepackage{tikz}
\usepackage{pgfplots}

\usetikzlibrary{3d,decorations.text,shapes.arrows,positioning,fit,backgrounds}

\tikzset{
}
\pgfplotsset{
    every axis/.style={
      xlabel style={
	      anchor=north,
	      font=\footnotesize},
      ylabel style={
	      anchor=north,
	      font=\footnotesize},
    }
}

\definecolor{col1}{RGB}{255,0,0}
\definecolor{col2}{RGB}{0,0,0}

\newcommand{\unknown}{x}
\newcommand{\txt}{t}
\newcommand{\realauthor}{\alpha}
\newcommand{\otherauthor}{\beta}
\newcommand{\txts}{T}
\newcommand{\txtscomp}[1]{\overline{#1}} 
\newcommand{\knowntxts}{\txts_\realauthor}

\newcommand{\threshold}{\delta}
\newcommand{\simfuncname}{s}
\newcommand{\simfunc}[2]{\simfuncname(#1,#2)}
\newcommand{\combinefunc}{C}
\newcommand{\trainset}{T_{train}}
\newcommand{\testset}{T_{test}}
\newcommand{\valset}{T_{val}}
\newcommand{\blburrow}{\textsc{Burrows}}
\newcommand{\blsvm}{\textsc{SVM}}

\newcommand{\powerset}[1]{\mathcal{P}\left(#1\right)}
\newcommand{\timefunc}[1]{\tau\left(#1\right)}

\newcommand{\Sim}{\textsc{Sim}}
\newcommand{\AV}{\textsc{AV}}

\newcommand{\embeddinglayer}{\textsc{Embd}}
\newcommand{\convAlayer}{\textsc{Conv8}}
\newcommand{\convBlayer}{\textsc{Conv4}}
\newcommand{\GMP}{\textsc{GMP}}
\newcommand{\Merge}{\textsc{Merge}}
\newcommand{\Dense}{\textsc{Dense}}

\newcommand{\FAR}{\mbox{FAR}}
\newcommand{\CR}{\mbox{CR}}
\newcommand{\TP}{\mbox{TP}}
\newcommand{\FP}{\mbox{FP}}
\newcommand{\TN}{\mbox{TN}}
\newcommand{\FN}{\mbox{FN}}

\newcommand{\secref}[1]{Section~\ref{sec:#1}}
\newcommand{\figref}[1]{Figure~\ref{fig:#1}}
\newcommand{\tabref}[1]{Table~\ref{tab:#1}}

\begin{document}
\title{Detecting Ghostwriters in High Schools}

\author{Magnus Stavngaard \hfill August S\o rensen \hfill Stephan Lorenzen$^{\dag}$\\
Niklas Hjuler \hspace{4mm} Stephen Alstrup
%
\thanks{Supported by the Innovation Fund Denmark through the Danish Center for Big Data Analytics Driven Innovation (DABAI). The authors would like to thank MaCom.}
%
\vspace{.3cm}\\
%
University of Copenhagen - Department of Computer Science \\
Universitetsparken 3, Copenhagen, Denmark\\
$\dag$ Corresponding author, e-mail: lorenzen@di.ku.dk
}


\maketitle

\begin{abstract}
Students hiring \emph{ghostwriters} to write their assignments is
an increasing problem in educational institutions all over the world, with companies selling these services as a product.
In this work, we develop automatic techniques with special focus on detecting such ghostwriting in high school assignments.
This is done by training deep neural networks on an unprecedented large amount of data
supplied by the Danish company MaCom, which covers 90\% of Danish high schools.
We achieve an accuracy of 0.875 and a AUC score of 0.947 on an evenly split data set.
\end{abstract}

\section{Introduction}
The number of Danish high school students using ghostwriters for their assignments has been
rising at an alarming rate due to
the emergence of several new online services,
allowing students to hire others to write their assignments\cite{tv2}.

We consider in this paper the problem of detecting such ghostwriting, or as it is more commonly known: \emph{authorship verification}.
Authorship verification is a common task in natural language processing \cite{stamatos2009,bagnall:2015,bartoli2015author}:
Given author $\realauthor$ with known texts $\txt\in\knowntxts$ and unknown text $\unknown$, determine whether $\realauthor$ is the author of $x$.
Often, a set of texts $\txtscomp{\knowntxts}=\txts\setminus\knowntxts$ ($\txts$ denoting the complete set of available texts) not written by $\realauthor$ is also available, which can be utilized as examples of different writing styles, when training a model. Note however, that $\txtscomp{\knowntxts}$ is unlikely to contain examples written by the true author of $\unknown$,
unlike in the related \emph{authorship identification} problem, in which the task is to determine the exact author of $\unknown$, given a set of candidate authors and their texts \cite{qian:2018,hansen2014}.


In this paper, we focus on the problem in high schools. We have access to a large data set
consisting of 130K Danish essays, written by more than 10K high school students\footnote{The data set is proprietary and not publicly available.}. Thus we have access to a lot of different authors, each with a large amount of text. 
We suggest a \emph{generalizing}
technique for authorship verification (as opposed to \emph{author specific} models);
using a Siamese network working at character level
(an approach inspired by \cite{qian:2018}),
writing style representations are learned and compared, in order to compute the style similarity between two texts.
Using the similarity measure provided by this network, $\unknown$ are compared to previous works $\txt\in\knowntxts$, and a final answer is given by a weighted combination of the individual similarities.
The data used is supplied by
MaCom, the company behind Lectio, the largest learning management system in Denmark.

Many previous approaches for authorship verification/identification are based on excessive feature selection \cite{Burrows02,stamatos2009}, but neural network approaches have also been considered, for instance
\cite{bagnall:2015} who utilize recurrent neural networks for identification.
Previous work on Danish high school essays have used author specific models for verification/identification \cite{hansen2014}, but this work is the first neural network based approach used on this data (and, to our knowledge, in this setting). 

\section{Method}
As mentioned, we solve the authorship verification problem in two steps.
First, we solve the problem of computing the writing style similarity between two texts by learning the similarity function $\simfuncname: \txts\times\txts \to [0,1]$ using a Siamese network (\secref{network}).
Second, we solve the authorship verification problem for author $\realauthor$ by combining similarities computed between the unknown text $\unknown$ and the known texts $\txt\in\knowntxts$. We consider several different ways to combine these similarities, based on their value and relevant meta data. (\secref{combine}).

\subsection{Network}\label{sec:network}
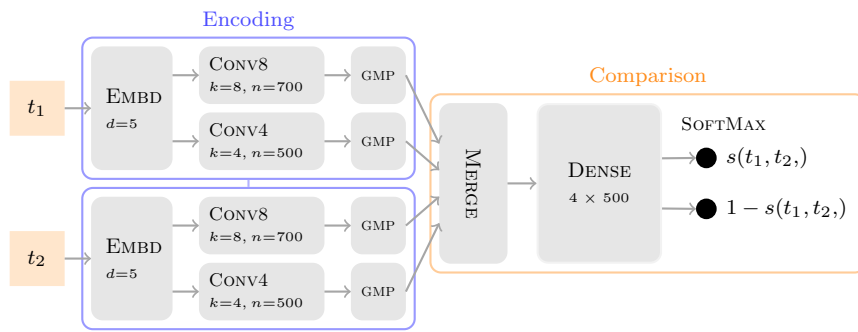
\begin{figure}[h!]
    \centering
    \begin{tikzpicture}[
  node distance = 10mm,
  every node/.style = {
    font=\tiny
  },
  einput/.style = {
    rectangle,
    draw=orange!20,
    font=\footnotesize,
    thick,
    fill=orange!20,
    minimum width=2em,
    minimum height=2em
  },
  eembd/.style = {
    rectangle,
    draw=gray!20,
    font=\footnotesize,
    thick,
    rounded corners,
    fill=gray!20,
    minimum width=3em,
    minimum height=4.6em
  },
  econv/.style = {
    rectangle,
    draw=gray!20,
    font=\scriptsize,
    thick,
    rounded corners,
    fill=gray!20,
    minimum width=4.5em,
    minimum height=2.1em
  },
  egmp/.style = {
    rectangle,
    draw=gray!20,
    thick,
    rounded corners,
    fill=gray!20,
    minimum width=2em,
    minimum height=2.1em
  },
  emerge/.style = {
    rectangle,
    draw=gray!20,
    thick,
    rounded corners,
    fill=gray!20,
    minimum width=2.5em,
    minimum height=6em
  },
  edense/.style = {
    rectangle,
    draw=gray!10,
    font=\footnotesize,
    thick,
    rounded corners,
    fill=gray!20,
    minimum width=4em,
    minimum height=6em
  },
  eout/.style = {
    circle,
    draw=black,
    thick,
    fill=black,
    inner sep=0pt,
    minimum width=0.7em,
    minimum height=0.7em
  },
  cedge/.style = {
    ->,
    draw=gray!70,
    thick
  }
]
  \node[einput] (input1) at (0,1) {$\txt_1$};
  \node[einput] (input2) at (0,-1) {$\txt_2$};
  \node[emerge] (merge) at (16.5em,0) {};
  \node[rotate=-90] at (16.5em,0) {\footnotesize \Merge};
  \foreach \X in {1,2} {
    \node[eembd, text width=2em]  (embd\X) [right = 1em of input\X .east] {\embeddinglayer\ \tiny $d$=$5$};
    \node[econv, text width=4em]  (conv\X a) [below right = 0.0em and 1em of embd\X .north east]
         {\convAlayer\ \ \ \ \ \tiny $k$=$8,n$=$700$};
    \node[econv, text width=4em]  (conv\X b) [above right = 0.0em and 1em of embd\X .south east]
         {\convBlayer\ \ \ \ \ \tiny $k$=$4,n$=$500$};
    \node[egmp]  (gmp\X a) [right = 1em of conv\X a.east] {\GMP};
    \node[egmp]  (gmp\X b) [right = 1em of conv\X b.east] {\GMP};
    \draw[cedge] (input\X) -> (embd\X);
    \draw[cedge] ([yshift=1.25em] embd\X .east) -> (conv\X a.west);
    \draw[cedge] ([yshift=-1.25em] embd\X .east) -> (conv\X b.west);
    \draw[cedge] (conv\X a) -> (gmp\X a);
    \draw[cedge] (conv\X b) -> (gmp\X b);
    
  }
  \draw[cedge] (gmp1a.east) -> ([yshift=1.5em] merge.west);
  \draw[cedge] (gmp1b.east) -> ([yshift=0.5em] merge.west);
  \draw[cedge] (gmp2a.east) -> ([yshift=-0.5em] merge.west);
  \draw[cedge] (gmp2b.east) -> ([yshift=-1.5em] merge.west);
  
  \node[edense, text width=4em, align=center] (dense) [right = 1.1em of merge.east] {\Dense\ \tiny $4\times 500$};
  
  \draw[cedge] (merge.east) -> (dense.west);

  \node[eout] (out1) [below right = 1.8em and 1.4em of dense.north east] {};
  \node[eout] (out2) [above right = 1.8em and 1.4em of dense.south east] {};
  \node[] (outl1) [right = 0.0em of out1.east] {\footnotesize $\simfunc{\txt_1,\txt_2}$};
  \node[] (outl2) [right = 0.0em of out2.east] {\footnotesize $1-\simfunc{\txt_1,\txt_2}$};
  \node[] [above right = 0.3em and -1.3em of out1.north] {\scriptsize \textsc{SoftMax}};
  
  \draw[cedge] ([yshift=0.95em] dense.east) -> (out1.west);
  \draw[cedge] ([yshift=-0.95em] dense.east) -> (out2.west);

  \begin{scope}[on background layer]
    \node[draw=blue!40,thick,rounded corners,fit=(embd1) (gmp1b)] (encode1) {};
    \node[draw=blue!40,thick,rounded corners,fit=(embd2) (gmp2b)] (encode2) {};
    \draw[thick,draw=blue!30] (encode1) -- (encode2);
    \node[blue!70] [above = -0.1em of encode1.north] {\footnotesize Encoding};
    \node[draw=orange!40,thick,rounded corners,fit=(merge) (outl2)] (comparison) {};
    \node[orange!90] [above = -0.1em of comparison.north] {\footnotesize Comparison};
  \end{scope}
  
\end{tikzpicture}
    \caption{Network architecture.}\label{fig:network}
\end{figure}

Several different architectures are considered, using different input channels (e.g. char, word, POS-tags), and evaluated on a validation set. The architecture of our best performing network is shown in \figref{network}.

The Siamese network can be considered in two parts: \emph{encoding}
and \emph{comparison}, the main idea being to learn an encoding of writing style, that the network is then able to distinguish. Our network uses only character level inputs.

The \textbf{encoding part} consists of a character embedding (\embeddinglayer), followed by two different convolutional layers: \convAlayer\ using kernel size $k=8$ and $n=700$ filters, and \convBlayer\ using $k=4$ and $n=500$. Each convolutional layer is followed by a global max pooling layer (\GMP). The weights of \embeddinglayer\ and \convAlayer/\convBlayer\ are shared between encoding $\txt_1$ and $\txt_2$.

In the \textbf{comparison part}, we first compute the absolute difference between the encodings in the \Merge\ layer. Afterwards, 4 dense layers with 500 neurons each are applied (\Dense), and finally, the output is normalized by use of a softmax layer with two outputs.

\subsection{Combining similarities}\label{sec:combine}
Having a good estimate of $\simfunc{\txt_1}{\txt_2}$ for any two texts, we consider different ways to combine these similarities, in order to give the final answer to an authorship verification query. More specifically, we consider functions $\combinefunc_\simfuncname : \powerset{\txts}\times \txts \to [0,1]$, such that, given $\unknown$ and $\knowntxts$,
we will answer the query positively (i.e. $\realauthor$ is the author of $\unknown$) if:
\begin{align*}
    \combinefunc_\simfuncname\left(\knowntxts,\unknown\right) \geq \threshold
\end{align*}
where $\threshold$ is a configurable threshold, which describes how likely we are to answer positively. In the experiments, we consider several different ways to combine similarities, for instance
using weighted sums, the min/max similarity or majority vote, while utilizing meta data such as time stamps and text length. From the experiments, we found that the optimal strategy was a weighted sum with weights decaying exponentially with time:
\begin{align}
    \combinefunc_\simfuncname\left(\knowntxts,\unknown\right) = \sum_{\txt\in\knowntxts} e^{-\lambda \timefunc{\txt}}\simfunc{\txt}{\unknown}
    \label{eq:combine}
\end{align}
where $\timefunc{\txt}$ denotes the time in months since $\txt$ was written, and $\lambda$ is a configurable parameter, which is determined experimentally.

\section{Experiment}
This section describes our experiments performed on the MaCom data.
\secref{data} will describe the preprocessing and partitioning of data.
Baselines will be described in \secref{baseline}. 
Finally, \secref{results} lists and discusses the final results.
We use accuracy, \emph{false accusation rate}, $\FAR = \FN/(\TN+\FN)$, and \emph{catch rate}, $\CR = \TN/(\TN+\FP)$ as performance metrics.

\subsection{Data}\label{sec:data}
The data
is partitioned into three sets: $\trainset$
used for training,
$\valset$ used 
for early stopping and selecting $\combinefunc_\simfuncname$, and
$\testset$
used only
for estimating the metrics of the
final models.
The three sets are author disjoint, meaning no author will appear in more than one of the sets.
In an effort to remove invalid data (blank hand-ins, etc.), we clean the data by filtering according to length (keeping texts with lengths between 400 and 30,000 characters). Furthermore, some texts were found to include author revealing information (such as name, address); hence we removed all proper pronouns from the texts, as well as the first 200 characters. Finally, authors with less than 5 texts were removed.

After cleaning, the data set contains a total of 131,095 Danish essays, written by 10095 authors, with an average 13.0 texts per author, and an average text length of 5894.8 characters.

For each data set, we construct two types of problem instances: \Sim\ and \AV, used for training the network and selecting the combination strategy respectively. The data set has no labelled ghostwriters, so we assume all authors to be correct\footnote{An undoubtedly false assumption, which will be discussed in \secref{results}}, and construct balanced (50/50) data sets as follows:

A \Sim\ instance simply consists of two texts $\txt_1,\txt_2$ and a label indicating whether the texts are by the same author. Positive samples are generated by using $\txt_1,\txt_2\in\knowntxts$, while negative samples are generated by using $\txt_1\in\knowntxts$ and $\txt_2\in\txtscomp{\knowntxts}$.
An \AV\ instance consists of a set of known texts $\knowntxts'$, an unknown text $\unknown$, and a label indicating whether $\realauthor$ is (positive) or is not (negative) the author of $\unknown$. Letting $\txt_{last}$ denote the most recent text of $\knowntxts$, samples are generated using $\knowntxts'=\knowntxts\setminus\{\txt_{last}\}$ with $\unknown=\txt_{last}$ for the positive sample, and $\unknown\in\txtscomp{\knowntxts}$ chosen at random for the negative sample.


\tabref{data} provides an overview of the data after partitioning and preprocessing.
\begin{table}[h!]
  \centering
  \begin{tabular}{|c|c|c|c|c|}
    \hline
    Data set    & \#authors & \#texts & \#\Sim & \#\AV \\
    \hline
    $\trainset$ & 5418        & 70432      & 934720  & 10836 \\
    $\valset$   & 989         & 12997      & 173536  & 1978  \\
    $\testset$  & 3688        & 47666      & 627744  & 7376  \\
    \hline
  \end{tabular}
  \caption{Data set overview.
  }\label{tab:data}
\end{table}

\subsection{Baselines}\label{sec:baseline}
We will compare our method to Burrows's Delta method and author specific SVMs:

Burrows's Delta method (\blburrow) \cite{Burrows02} is a method for authorship identification based on the $l_1$-distance between the $z$-scores of word frequencies in $\unknown$ and in the corpus for each of the candidate authors $\otherauthor_1,...,\otherauthor_k$. We adapt it for verification by sampling a set of 'wrong' authors, $\otherauthor_2,...\otherauthor_k$, and querying with $\unknown$ and $\otherauthor_1 = \realauthor, \otherauthor_2,...,\otherauthor_k$. answering positively, if $\unknown$ is attributed to $\realauthor$. The top 150 word frequencies are considered.
The optimal $k$ is determined using $\trainset$.

An author specific SVM \cite{hansen2014,stamatos2009} is trained for each author in order to recognize $\knowntxts$ from $\txtscomp{\knowntxts}$.
Hyper parameters and features are selected using cross validation. Forward feature selection is used, considering char, word and POS-tag $n$-grams for varying $n$.
The SVM will be trained on a balanced set, meaning that only a limited amount of data is available for each SVM.
However, they have previously been shown to work well in this data set \cite{hansen2014}.

\subsection{Results}\label{sec:results}
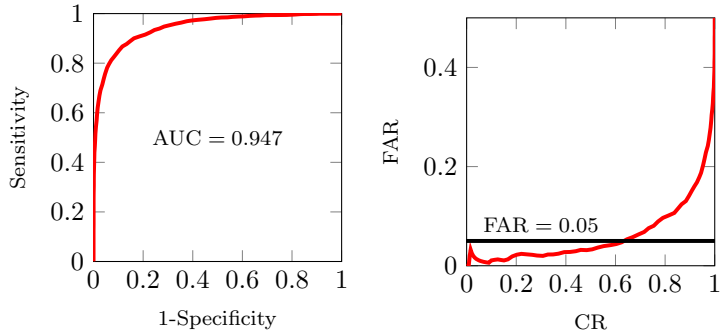
\begin{figure}
\centering
\begin{tikzpicture}
	\begin{axis}[
			width=0.4\linewidth,
			height=0.4\linewidth,
			xmin=0,xmax=1,
			ymin=0,ymax=1,
			xlabel={1-Specificity},
			ylabel={Sensitivity},
			]
		\addplot[mark=none, color=col1, line width=1.5pt] table [x=FPR, y=TPR, col sep=semicolon] {data/plot.csv};
		\node[] at (axis cs: 0.5,0.5) {\footnotesize $\mbox{AUC}=0.947$};
	\end{axis}
\end{tikzpicture}
\begin{tikzpicture}
	\begin{axis}[
			width=0.4\linewidth,
			height=0.4\linewidth,
			xmin=0,xmax=1,
			ymin=0,ymax=0.5,
			xlabel={CR},
			ylabel={FAR},
			]
		\addplot[mark=none, color=col1, line width=1.5pt] table [x=TNR, y=FAR, col sep=semicolon] {data/plot.csv};
  		\addplot[mark=none, color=col2, samples=2, line width=1.5pt] {0.05};
		\node[anchor=south] at (axis cs: 0.3,0.05) {\footnotesize $\FAR = 0.05$};
	\end{axis}
\end{tikzpicture}
\caption{ROC (left) and plot of false accusation rate/catch rate (right) on $\testset$.}
\label{fig:ROC}
\end{figure}

Methods were trained and validated on $\trainset$ and $\valset$. For $\blburrow$, we found $k=4$ to give the best results, while the parameters $C=10,\gamma=10^3$ were found optimal for the RBF kernel $\blsvm$. The optimal combination strategy $\combinefunc_\simfuncname$ was found to be exponentially decaying weights (see \eqref{eq:combine}) with $\lambda = 0.1 $. Furthermore, $\threshold=0.57$ was found to be optimal. Using these parameters, the baselines and our method were evaluated on $\testset$; \tabref{results} presents the results, while \figref{ROC} shows the ROC/AUC and a plot of false accusation/catch rate for our method.
As it can be seen, our method clearly outperforms the baselines, on all metrics.
\begin{table}[h!]
  \centering
  \begin{tabular}{|c|c|c|c|}
    \hline
    Method      & Accuracy   & \FAR      & \CR    \\
    \hline
    $\blburrow$ & 0.677      & 0.357     & 0.806  \\
    $\blsvm$    & 0.720      & 0.266     & 0.689  \\
    Our method  & 0.875      & 0.141     & 0.896  \\
    \hline
  \end{tabular}
  \caption{Results obtained on $\testset$}\label{tab:results}
\end{table}

The false accusation rate is especially important considering the use case:
when trying to detect ghostwriting in high schools, making false accusation can be especially devastating, as students found guilty of cheating could risk severe punishment and maybe even be expelled.
Using this metric, our method performs very well, as illustrated in \figref{ROC} (right), a fairly low \FAR\ can be obtained, while still catching a lot of ghostwriters.
Optimizing the method on $\valset$ while restricting $\FAR < 0.1$, we achieved an accuracy of $0.864$, $\FAR=0.106$ and $\CR=0.825$ on $\testset$ (with exponential weighting and parameter $\lambda = 0.16$).
However, even if these results are promising, the system should only be used as a warning system for the teacher, who should always have the final say.

An interesting aspect to note about the combination strategy $\combinefunc_\simfuncname$, is
that it takes time into account with $\lambda =0.1$,
weighing recent assignments more than older ones.
Since
$\timefunc{t}$ measures in months, this means that a
recent
assignment gets
$e^{12 \cdot 0.1} \approx 3.3$ times the weight of
a one year old assignment. This corresponds well with the idea that high school students writing style changes over time,
as also observed in \cite{hansen2014}.

When looking at the low false accusation rates of \figref{ROC} (right), one have to consider two things before translating them into practice:
a)
$\testset$ is
balanced,
while in reality much less than half of assignments are written by a ghostwriter,
and b)
ghostwriting does happen, also in our data set, and thus most likely some of our labels are wrong.
A possible remedy for the
second point
could be
to adjust $\FN$ to $\FN-\frac{\TN}{\TN+\FP} \gamma\mbox{T}$ (where $\gamma$ is the estimated fraction of ghostwriters and $\mbox{T} = \TP+\FN$), and similar for $\TP$,
under the assumption that a negative sample and a corrupted positive sample are indistinguishable. Adjusting for this would obviously lead to improved accuracy and false accusation rate, but
requires a good estimate of $\gamma$.

\section{Conclusion}
We achieved an accuracy of 0.875, with a false accusation rate of 0.141 and a catch rate of 0.896. We show how false accusation rate can be improved at the cost of catch rate and accuracy. Results are good enough for practical use,
and even with a slightly lower catch rate, the system is still expected to have a preventive effect.
However, one has to keep in mind that, in practice, the data set is not 50/50 balanced, which obviously will affect the results. Making a split imitating the real world is hard for two reasons:
one needs a good approximation of the actual fraction of ghostwriters,
and even if this fraction is known,
the number of corrupt labels would be approximately the same as the number of negatives, making it impossible to beat a false accusation rate of 0.5, even for a perfect classifier.
Finding a clean data set or establishing ground truth would alleviate these problems, and could be interesting prospects for future work.

Another interesting direction
is to analyze writing style changes over time more in depth, motivated by
the chosen combination strategy and preliminary experiments,
which show how two texts written within a shorter time span have higher similarity on average.



\begin{footnotesize}

\bibliographystyle{unsrt}
\bibliography{ref.bib}

\end{footnotesize}


\end{document}